# Comprehensive Autonomous Vehicle Optimal Routing With Dynamic Heuristics

Ragav. V[a], Jesher Joshua[b], Syed Ibrahim S P[C*]

[a]ragav.v2021@vitstudent.ac.in, [b]jesherjoshua.m2021@vitstudent.ac.in, [c]syedibrahim.sp@vit.ac.in, Vellore Institute of Technology, Chennai Campus, India

## Abstract

Auto manufacturers and research groups are working on autonomous driving for long period and achieved significant progress. Autonomous vehicles (AV) are expected to transform road traffic reduction from current conditions, avoiding accidents and congestion. As the implementation of an autonomous vehicle ecosystem includes complex automotive technology, ethics, passenger behaviour, traffic management policies and liability etc., the maturity of AV solutions are still evolving. The proposed model to improve AV user experience, uses a hybrid AV Network of multiple connected autonomous vehicles which communicate with each other in an environment shared by human driven vehicles. The proposed Optimal AV Network (OAVN) solution provides better coordination and optimization of autonomous vehicles, improved Transportation efficiency, improved passenger comfort and safety, real-time dynamic adaption of traffic & road conditions along with improved in-cabin assistance with inputs from various sensors. The true optimal solution for this problem, is to devise an automated guidance system for vehicles in an AV network, to reach destinations in best possible routes along with passenger comfort and safety. A custom informed search model is proposed along with other heuristic goals for better user experience. The results are analysed and compared to evaluate the effectiveness of the solution and identify gaps and future enhancements.

## 1.0 Introduction

The concept of autonomous driving has been in talks for more than seven decades. In spite of several attempts by engineers to give a shape to this idea, failed to succeed. From the first attempt by Ernst Dickmanns in 80's called VaMP[1], which was able to navigate in a controlled environment through cameras and sensors, there are umpteen attempts made.

During late 90's DARPA Grand Challenge[2] which was held to develop autonomous vehicles for military purposes, triggered the design of many autonomous vehicle technologies, including radar and LiDAR sensors. Google's self-driving car project, Waymo, came in 2009, and Tesla introduced its Autopilot in 2015.

To completely bring the benefits of AV technology, all vehicles need to be connected both to one another and to roadside systems (traffic light systems, traffic monitoring, emergency or maintenance services). These connections in the network need to be in real-time with low latency and high reliability leveraging benefits of modern networks like 5G services.

Typically AV connects to the network for V2V (Vehicle to Vehicle) and vehicle to network (V2N) communication through dedicated network to exchange information  AV networks can be rated in 5 stages, like Assisted Operations, Partial autonomous, conditional, High and fully autonomous stages.

As AV network size increases the operational expenses to manage them, predictive maintenance and optimal routing needs increases. The proposed model collates data from various sensors, apply Machine Learning(ML) models and optimize through hybrid path routing algorithms for better navigation and traffic avoidance

## 2.0 Objectives and Assumptions

When the AV networks mature with increased adaptability, the need for intra-vehicle communication and optimizing their routes to meet the goals and constraints are necessary. For this a Coordinated Autonomous Vehicle Decision Making with Network Heuristics optimization approach is proposed, which learns & adapts to the environment.

i. **Goal:** The goal of proposed model is to 'AV to reach optimally from source to the destination in a network with environment & user-controlled heuristics, meeting defined optimality conditions.

ii. **Optimality Conditions**: Minimum distance (minimum energy consumption), Minimum time to destination, Maximum Comfort & Safety

iii. **Assumptions of proposed Solution:**

The model assumes that all Autonomous vehicles are connected and controlled by the same network.

Autonomous vehicles are in a mixed environment with human-driven vehicles

All the basic AV manoeuvring features are matured and available such as object detection, obstacle avoidance, network communication etc, while the solution focuses on the macro network optimal navigation.

## 3.0 AV Network Environment

Autonomous vehicles need increasingly sophisticated systems, with high performance computers and an large number of advanced driver assistance system sensors, such as HR cameras, LIDAR ( Light Detection and Ranging ) and noise sensors.

These eco-systems generating large amounts of data, which demands high-speed data nodes, links, swtiches and assemblies. The inside of the AV is like an information highway with data streams flowing in different directions managed for better comfort and safety. Automotive OEM manufacturers to follow the network connectivity standards. The network support depends on the need like infotainment to ADAS (Driver Assistance) with bandwidth need to 1 to 25 GBPS range.

V2X, (vehicle-to-everything) interfacing technology permits vehicles to directly communicate with each other, roadside infra / users to deliver road safety, traffic management, smart mobility & sustainability of environment. Both sensor & radio-based interfaces are needed for V2X[3].

Vehicle to Network (V2N) aims to send information among vehicles through high-bandwidth/reliability & low-latency network eco system. V2N systems integrate vehicles with data centres consonantly for stable information exchange. V2N network also connects vehicles with other vehicles in the network. Though popular apps can provide traffic updates and navigation, they have limitations. Vehicles share more detailed information about their position, status, speed, direction, intersection movement assist (IMA) and Left turn assist (LTA) to improve safety. V2N can also prompt pedestrian devices (V2P) to avoid possible accidents.

Figure.1 below details about the Autonomous vehicle network environment assumed.

i) The AV cloud infra stores HD maps, multimedia streams from vehicles and live traffic data as V2N update. This also provides computing infrastructure to execute Machine language and routing algorithms proposed.

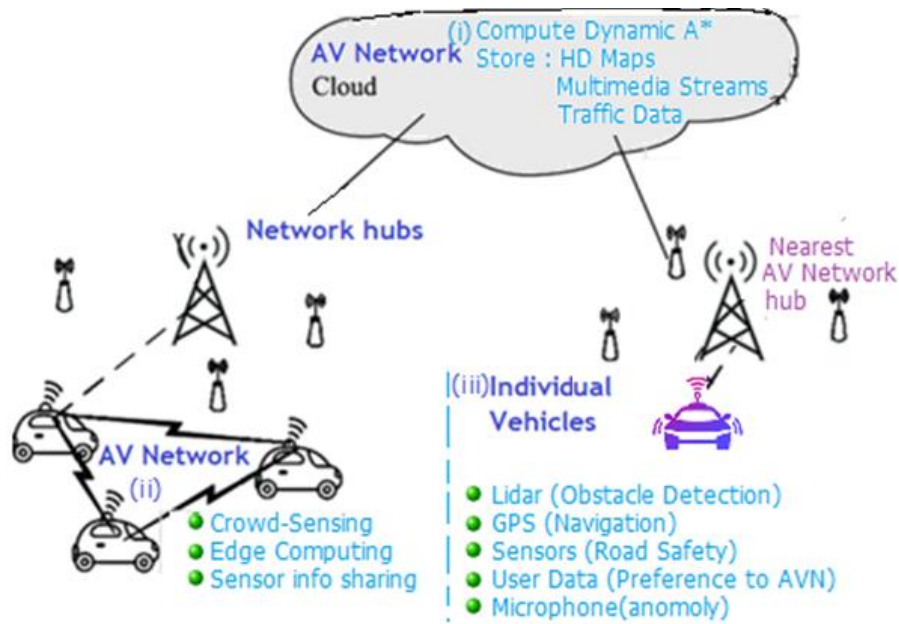

Fig.1 AV Network connectivity, storage and attributes

ii) The AV network consisting of set of cars connecting to a hub of the network is responsible for sharing crowd sensing data of constituent vehicles with local computing capabilities (Edge distributed) and share the sensor data to the AV cloud.

iii) This section details on various sensors available on individual AV vehicles like GPS, Lidar, HD Cameras, ultrasound sensors etc. which collate in-cabin and outside data collected, shared to the cloud periodically. Details of figure.1(iii) is elaborated in figure-2 below, with complete flow.

## 4. Proposed Optimization model

### 4.1 Model Component and interfaces

The AV optimal routing model is based on sensing (perception), understanding localized scenarios, dynamic path planning & guided navigation control. Among these Path planning is the complex step to find of a collision-free path in a given constrained real-time environment[4]. It is a non-deterministic polynomial-time hard problem to solve [5]. Different sensors, and interfaces used are detailed in Figure-2, along with device details, purpose and algorithms used for different tasks. The AV sensors are categorized based on their location such as in-cabin, outside AV and inside AV-machinery loaded sensors. The inside vehicle sensors used are microphone, HD cameras, smoke detector, IR (for user cabin temperature control) etc. Outside sensors used are Lidar, Ultrasonic sensors, cameras etc. Inside AV sensors include microphone (for wear and tear detection using anomaly detection) and sensors for engine health monitoring.

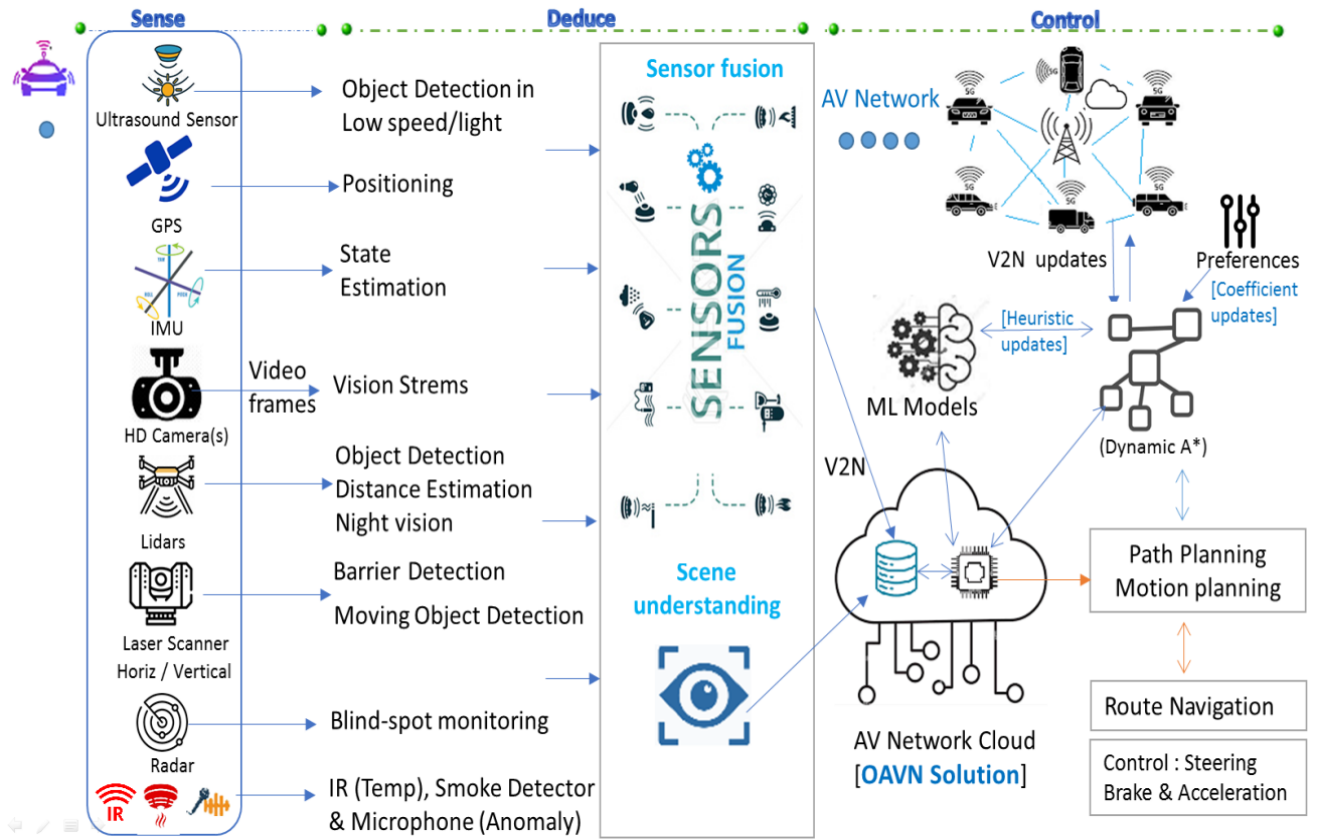

Figure. 2 Dynamic A* Model environment, interfaces & flow

**4.2 AV Sensors, functions and ML algorithms**

| Sensors | Devices | Location | Purpose | ML Algorithms |
|---|---|---|---|---|
| Microphone | MEMS / Knowles SPM0687LR5H-1 | Inside AV | Detect abnormal noise patterns / problems | CNN / RNN |
| Camera | Raspberry Pi Camera Module V2 | Inside AV | Passenger behaviour, dangerous objects, Children safety | CNN / SVM |
| Smoke detector | Optical - Honeywell 5808W3 | Inside AV | detect smoke or other patterns indicating a fire | Random forest |
| IR Sensor | Melexis MLX90614 | Inside AV | Temperature / Climate prediction & control | MLP |
| Lidar | Velodyne Puck | Outside AV | Obstacle detection and avoidance | CNN / RNN |
| Ultrasonic sensor | HC-SR04 | Outside AV | Prevent close proximity accidents with distance to nearby objects | Linear regression/ Decision tree |
| HD Camera | FLIR Blackfly - BFS-U3-50S5C | Outside AV | Object detection &, Obstacle avoidance | CNN / SVM |
| Microphone | MEMS / Knowles SPM0687LR5H-1 | AV Machinery | Vehicle wear and tear detection - Anomaly detection | 1-class SVM / Isolation forest |
| Sensor (Engine) | TE Connectivity MS5839-02BA01 | AV Machinery | Monitor the health of the engine and detect signs of problems | Random forest or a support vector regression (SVR) |

Table 1 AV sensors, functions and ML algorithms

All the sensor data and video streams are collated by Sensor fusion methods and scene / environment visualization is deduced. The single vehicle sensor streams and fusion data is transmitted to AV cloud storage with vehicle identification. Similar information is collected from multiple vehicles in the Optimal AV Network (OAVN) proposed as a comprehensive solution.

The multi-vehicle sensor data is updated as V2N streams to the cloud. Different machine learning (ML) model proposed as in Table-1, are executed in cloud computing infrastructure along with minor pre-processing done at edge nodes. The system/user/passenger preferences on comfort parameters are also fed to the Network.

The ML outputs are fed as heuristic parameter updates along with user preferences as co-efficient of dynamic A* algorithm for optimal route search. The dynamic A* algorithm getting updated node data continuously from environment direct the component vehicles in optimal route to avoid collision and traffic in better route for comfort and safety.

### 4.3 The Dynamic A* Network route optimization:

To select an optimal route from the source to destination, various search algorithms such as uniform cost search (UCS)[6], greedy search (best first search), A* search were compared and analysed. Among the algorithms A* performed better[7-8], compared to other on similar environment. The proposed dynamic A* search model works to meet the varying constraints and responses from the environment[9]. The changes are represented by the heuristic function at each node[10]. The ML model outputs define the changes to heuristic values and its coefficients, which are optimized by the dynamic A*

An optimal search solution using the A* Search is chosen, with variable heuristics (Distance, Time, Comfort, Safety) parameters. This model combines the results from computer vision through camera frames which detects obstacles, road conditions and in-cabin safety aspects through ML Model learning is proposed.

Dynamic A* approach:

Let $f(n)$= the estimated cost of the solution

$h_1(n)$ = heuristic value with respect to time taken to reach the destination.

$h_2(n)$ = heuristic value with respect to comfort (road condition/traffic)

$h_3(n)$ = heuristic value for safety (constant & not changed for any condition)

Thus, $f(n)$ is calculated by:

$$f(n) = g(n) + h_1(n) + h_2(n) + h_3(n) \qquad (1)$$

This provides optimal cost function for the normal heuristics' conditions assumed.

This approach modifies the equation (1) with variable coefficients which are a result of Combining current heuristics cost, with inputs from the computer vision & in-cabin predicted model outputs, a new cost value is found.

$$f_1(n) = w_1\, g(n) + w_2\, h_1(n) + w_3\, h_2(n) + w_4\, h_3(n) \qquad (2)$$

where, w1 , w2 and w3 are the coefficients that determine the weight of each heuristic.

For example, based on the sensor output, which identifies quality of the road and the past experience of other vehicles travelled on same path earlier, the model analyses the comfort on travel and alters the coefficients of the heuristics (weights) appropriately.

The value of the heuristic function of each node is calculated by the ML models dynamically and fed to the search algorithm which returns the optimal route. The dynamic A* algorithm pseudocode is described below:

```
Dyn_A_star (start, goal, w1, w2)
    create open set and closed set
    add start node to open set with g(start) = 0, and f(start) = w1 * h1(start) + w2 * h2(start) + w3 * h3(start)
    while open set is not empty do
        current_node = node in open set with lowest f value
        remove current_node from open set
        add current_node to closed set
        if current_node is goal then,    return path from start to goal
        for each successor of current_node do
            if successor is in closed set then,    continue
            tentative_g_score = g(current_node) + cost(current_node, successor)
            if successor is not in open set or tentative_g_score < g(successor) then
                set g(successor) = tentative_g_score
                set h1(successor) = heuristic_cost_based_on_time(successor)
                set h2(successor) = heuristic_cost_based_on_comfort(successor)
                set h3(successor) = heuristic_cost_based_on_safety(successor)
                set f(successor) = g(successor) + w1 * h1(successor) + w2 * h2(successor) + w3 * h3(successor)
                if successor is not in open set then,    add successor to open set
        update heuristics based on current state of the search
    end while
    return failure
end function
```

### 4.4 Machine learning AV Models:

As detailed above, ML outcomes determine real-time heuristic values of above dynamic A* model. Assisted ML Algorithms used for various tasks like Computer vision (Object Detection), In-Cabin Assistance (ML), Comfort (smoother roads and lesser traffic), preventive maintenance (with on machinery noises) are made mature with, repeated training and validation of data collected from different sensors The fusion modules collates different sensor outputs to understand the dynamically changing on-road scenarios.

For example and outside AV HD camera (FLIR Blackfly - BFS-U3-50S5C) is used for Object detection &, Obstacle avoidance  A machine learning model such as a convolutional neural network (CNN) or a support vector machine (SVM)[11] can be trained to analyse video/image streams, detect single / multiple objects and identify patterns that may indicate an obstacle as described below Pseudocode :

```
Collect and preprocess Camera stream data
Vision_data = collect_Vision_data()
processed_data = preprocess_vision_data(vision_data)
Train model
model = train_vision_model(processed_data)
Make predictions on new data
new_image_data = collect_new_vision_data()
processed_new_data = preprocess_vision_data(new_vision_data)
predictions = model.predict(processed_new_data)
```

The minimal bounding box algorithm, to detect objects orthogonal with image is stated as:

```
Till count of boxes reaches max_count
  make first guess of 2 coordinates
  till No. of guesses = max_guess / matching criteria met
    evaluate guess
    store guess & guess results
    improvise on guess based on results and
        injected random values,
        exclude : locations already passed
    if some transitional criteria is met
      change the guess, estimation, and improvise
          suitable for criteria match
  if no guess matched conditions
    break
```

A typical SSD300 object detector with Resnet-50 convoluted neural network (CNN) backbone is used for vehicle objects detection from a colour compensated image stream/frames[12]. The images are extracted and trained by a typical neural network with below listed parameters in table 2 and the ANN classifies the test images into multiple objects. The optimal ANN hyper parameters are listed in below table -2 are used as end stage classifier.

| ANN Parameter | Model Values |
| --- | --- |
| Hidden Layers | 50 |
| Node/hidden layer | 4 |
| Activation | SoftMax |
| Epoch (cycles) | 500 |
| Regularization | Lasso Reg. |
| Momentum | 0.18 |
| Gradient Descent | ADAM (Adaptive Momentum) |
| Learning rate | 0.045 |
| Error Tolerance | 0.0013 |
| Bias | -1, 0, +1 |

Table 2 : Optimal ANN parameters of end state neural classifier parsing ResNet-50 metrics

The SSD300[13] outputs from camera feeds of single vehicles are used to transmit data to OAVN cloud and used to deduce and update traffic clusters on a particular node of dynamic A* model. The same image analysis can derive information on road conditions, pedestrian movements, road signals etc., which are pre-processed locally and updated to central cloud AV network infrastructure.

## 5.0 Results analysis

In above sections the proposed several widely used path planning algorithms and their variants, categorized based on the static and dynamic characteristics it can handle and its adaptability. These result comparisons are listed in table-3 below. The Optimality & Adaptability score calculated from the scenarios, the model deduce and act correctly, with the total scenarios validated.

| Algorithm | Static Constraints | Dynamic Constraints | Optimality & Adaptability score |
|---|---|---|---|
| Dijkstra/ UCS, Greedy search | ✓ | | 0.51 to 0.59 |
| A* | ✓ | | 0.65 |
| Slime mould based optimization[14] | ✓ | | 0.58 |
| Rapidly-Exploring Random Trees[15] | | ✓ | 0.69 |
| A* with dynamic ML heuristics | | ✓ | 0.77 |

Table 3 : Result comparison different path finding algorithms

From the above analysis we find that the accuracy and relevance of proposed dynamic heuristic A* algorithm performing better with improved adaptability to real-time scenarios and environmental conditions.

## 6.0 Conclusion and Enhancements

In the forthcoming era, AVs will become an vital part of modern transport ecosystem[16]. The development of autonomous vehicle networks has also raised concerns about job loss, cybersecurity, and ethical dilemmas. Despite these challenges, the technology continues to advance, and it is expected to transform the transportation industry in the coming years. Perception & representation to apt models, of the real-life scenarios remains the highest challenge for reliable and smoother auto driving with safety measures. The proposed model stores the environmental data from the OAVN to the cloud and controls the dynamic heuristics with user-controlled coefficients and found to perform better with real-time conditions.

Future enhancements would be implementing a prototype on controlled simulation and strengthening the model, with the stabilized model, develop real world applications such as driving to nearest charging station, emergency handling (drive to nearest hospitals/ Secured zones), enhancing a hybrid environment with human driven & AV vehicles, network data analytics to learn from real world scenarios with improved system security and integrity.